\documentclass{article}

\usepackage{arxiv}

\usepackage[utf8]{inputenc} 
\usepackage[T1]{fontenc}    
\usepackage{hyperref}       
\usepackage{url}            
\usepackage{booktabs}       
\usepackage{amsfonts}       
\usepackage{amsmath}        
\usepackage{nicefrac}       
\usepackage{microtype}      
\usepackage{graphicx}
\usepackage{doi}

\usepackage{float}
\usepackage{placeins}
\usepackage{caption}

\title{What Does Debiasing Really Remove? \\
A Geometric Study of PCA-Based Gender Debiasing in Word Embeddings}

\author{
Alexey Kresin$^{1}$ \\
\texttt{ak26@hood.edu}
\and
Tchifou M. Dieffi$^{1}$ \\
\texttt{mdieffi@gmail.com}
\and
Tomer Caspi$^{2}$ \\
\texttt{tomer279@gmail.com}
\\[0.5em]
$^{1}$ Department of Computer Science, Hood College, Frederick, MD, United States \\
$^{2}$ Department of Mathematics, Ben-Gurion University of the Negev, Be'er Sheva, Israel
}

\hypersetup{
pdftitle={What Does Debiasing Really Remove? A Geometric Study of PCA-Based Gender Debiasing in Word Embeddings},
pdfauthor={Alexey Kresin, Tchifou M. Dieffi, Tomer Caspi},
pdfkeywords={word embeddings, debiasing, PCA, gender bias, WEAT, semantic geometry},
}

\begin{document}
\maketitle

\begin{abstract}
Debiasing methods based on principal component analysis (PCA) are broadly used to reduce gender bias in word embeddings used in LLMs, yet it remains unclear what aspects of bias they actually remove and how destructive this process is.
These methods are based on the understanding that bias resides in a low-dimensional subspace, with the assumption that most of it can be captured by a few principal components. In this work, we conduct a systematic geometric analysis of PCA-based gender debiasing and investigate what is actually removed from the embedding space.

Our experiments across multiple embeddings show that direct gender bias is primarily concentrated in the first principal component, supporting the low-rank bias hypothesis. However, associative bias measured by WEAT does not align with these principal directions and is instead spread across multiple embedding dimensions. Furthermore, as expected, we demonstrate that removing an increasing number of principal components leads to a consistent degradation of the embedding geometry, affecting semantic structure and vector relationships.

These results reveal that PCA-based debiasing operates as a trade-off: while it effectively reduces certain forms of direct bias, it fails to eliminate distributed associations and introduces geometric distortion. Moreover, there is no universal optimal level of debiasing, as the balance between bias reduction and semantic preservation depends on the chosen metric and embedding.

Overall, our findings suggest that bias in word embeddings is not purely low-rank and that simple subspace removal methods may be insufficient for comprehensive debiasing.
\end{abstract}


\section{Introduction}

Word embeddings play a main role in natural language processing and are widely used in large language models (LLMs). They provide a way to represent words as vectors, enabling systems to capture the semantic meanings and contextual relationships of words. Yet it has been widely observed that word embeddings also inherit and amplify societal biases from real world data, including gender bias. For example, certain professions may become disproportionately associated with one gender, which can lead to biased outcomes in real-world, such as favoring male candidates for leadership or management roles during hiring.

To address this issue, several debiasing methods are widely used, most of them rely on  principal component analysis (PCA). Those methods based on the assumption that bias lives primarily in a low-dimensional subspace. The goal of those methods is to identify and remove these directions to reduce bias while keeping semantic relationships.

Despite their effectiveness, the geometric mechanisms underlying PCA-based debiasing remain poorly understood. In particular, it is unclear what portion of bias is actually removed, how bias is distributed across the embedding space, and to what extent debiasing distorts the geometry that encodes semantic relationships.

In this work, we take a geometric and spectral perspective on PCA-based gender debiasing. We analyze the structure of the gender subspace using PCA and study the effects of removing principal components on both bias and embedding geometry. Our analysis distinguishes between \textit{direct bias}, which aligns with specific directions, and \textit{associative bias}, which is more distributed and is measured using WEAT.

Our experiments across multiple embeddings show that direct gender bias is highly concentrated in the leading principal component, providing strong empirical support for the low-rank bias hypothesis. This is further confirmed by the explained variance spectrum, which reveals that most gender-related variation is captured by a small number of principal components. However, associative bias does not align with these dominant directions and remains distributed across the embedding space. 

Notably removing more principal components consistently degrades embedding geometry. This reveals a key trade-off in PCA-based debiasing: reducing direct bias comes at the cost of semantic distortion, while associative bias remains, highlighting the limitations of current methods.

\section{Contributions}

This work provides a systematic geometric analysis of PCA-based gender debiasing in static word embeddings. Our main contributions are as follows:

\begin{itemize}

\item \textbf{Geometric characterization of debiasing.}  
We study PCA-based debiasing from a geometric perspective and analyze how removing principal components affects both bias and the structure of the embedding space.

\item \textbf{Distinction between direct and associative bias.}  
We show that direct bias is highly concentrated in the leading principal component, while associative bias, measured using WEAT, is distributed across multiple dimensions and does not align with the same principal directions.

\item \textbf{Single-PC vs. multi-PC analysis.}  
We introduce a systematic comparison between cumulative component removal and single-PC ablation, demonstrating that the first principal component dominates direct bias reduction, whereas geometric effects remain distributed across components.

\item \textbf{Cross-embedding validation.}  
We verify that these behaviors are consistent across GloVe, Word2Vec, and FastText, suggesting that the low-rank structure of direct bias is a general property of static word embeddings.

\item \textbf{Trade-off between debiasing and semantic preservation.}  
We show that removing additional components leads to increasing geometric distortion, revealing a fundamental trade-off between bias reduction and preservation of semantic structure.

\end{itemize}

\section{Methodology}

\subsection{Word Embeddings}

We performed our experiments using the following static word embeddings: GloVe, Word2Vec, and FastText. Words in these word embeddings are represented as dense vectors $v \in \mathbb{R}^d$, where semantic relationships between words are reflected in geometric properties such as distances and directions.

Prior to analysis all embeddings were normalized to ensure consistency across models and also to make geometric comparisons more meaningful. Formally, each word vector $v$ is normalized as:
\begin{equation}
\hat{v} = \frac{v}{\|v\|}
\end{equation}

\subsection{Bias Subspace and PCA}

To isolate the geometric directions capturing gender-related variation, we construct a set of $n$ definitional word pairs (e.g., \textit{he–she}, \textit{man–woman}) . For each pair $(w_i^{(m)}, w_i^{(f)})$, we compute the difference vector between their respective normalized embeddings:

\begin{equation}
d_i = v(w_i^{(m)}) - v(w_i^{(f)}) \quad \text{for } i = 1, 2, \dots, n \label{eq:diff_vec}
\end{equation}

These difference vectors serve as empirical indicators for the gender variation present in the embedding space. To formally define the gender subspace, we construct the difference matrix $D \in \mathbb{R}^{d \times n}$, whose columns correspond to these difference vectors:

\begin{equation}
D = [d_1, d_2, \dots, d_n] \label{eq:matrix_D}
\end{equation}

Following standard subspace identification frameworks \cite{bolukbasi2016}, we assume that the difference vectors are zero-mean centered across pairs. We define the sample covariance matrix $\Sigma \in \mathbb{R}^{d \times d}$ of the bias directions directly from $D$ as :

\begin{equation}
\Sigma = \frac{1}{n} D D^\top \label{eq:covariance}
\end{equation}

We then apply Principal Component Analysis (PCA) to $\Sigma$ to extract an orthogonal basis $\{u_1, u_2, \dots, u_k\}$ that captures the directions of maximum variance . The leading principal component $u_1$ is computed sequentially by solving the optimization problem :

\begin{equation}
u_1 = \arg\max_{\|u\|_2 = 1} u^\top \Sigma u \label{eq:optimization}
\end{equation}

Subsequent components are obtained under the same variance-maximization objective constrained by orthogonality to all principal directions previously determined. Under the low-rank bias hypothesis, the dominant semantic direction of gender bias is expected to be captured entirely within the subspace spanned by the first few leading principal components where variance is highest .

\subsection{Debiasing via Component Removal}

We study debiasing by progressively removing principal components associated with the bias subspace. Given a word vector $v$, we project it onto the identified principal components and subtract this projection from the original vector.

For removal of the first $k$ principal components, the debiased vector $v'$ is computed as:
\begin{equation}
v' = v - \sum_{i=1}^{k} (v \cdot u_i) u_i
\end{equation}

We consider two settings:

\begin{itemize}
    \item \textbf{Single-PC removal:} removing only the first principal component ($k=1$).
    \item \textbf{Multi-PC removal:} removing the first $k$ principal components, where $k$ varies across experiments.
\end{itemize}

This allows us to analyze how removing increasing amounts of the bias subspace affects both bias and the overall geometry of the embedding space.

\subsection{Evaluation Metrics}

To evaluate the effects of debiasing, we consider multiple complementary metrics:

\paragraph{Direct Bias.}
We measure direct bias by projecting word vectors onto the gender direction. Given a gender direction $g$, the direct bias for a word $w$ is defined as:

\begin{equation}
\mathrm{DB}(w) = |v(w) \cdot g|
\end{equation}

We compute the average direct bias over a predefined set of gender-neutral words.

\paragraph{Associative Bias (WEAT).}
We use the Word Embedding Association Test (WEAT) to measure associative bias. Given two target sets $X, Y$ and attribute sets $A, B$, the association of a word $w$ is:

\begin{equation}
s(w, A, B) = 
\frac{1}{|A|} \sum_{a \in A} \cos(v(w), v(a))
- 
\frac{1}{|B|} \sum_{b \in B} \cos(v(w), v(b))
\end{equation}

The WEAT score is then computed as the difference between associations of the two target sets.

\paragraph{Mean Vector Displacement.}
To quantify geometric distortion, we compute the average displacement between original and debiased vectors:

\begin{equation}
\mathrm{MVD} = \frac{1}{N} \sum_{i=1}^{N} \|v_i - v_i'\|
\end{equation}

\paragraph{Neighbor Stability.}
We analyze the stability of nearest neighbors for each word before and after debiasing. Let $\mathcal{N}_k(w)$ denote the set of $k$ nearest neighbors of $w$. Stability is measured as:

\begin{equation}
\mathrm{NS}(w) = \frac{\left| \mathcal{N}_k^{\mathrm{before}}(w) \cap \mathcal{N}_k^{\mathrm{after}}(w) \right|}{k}
\end{equation}

Changes in nearest neighbors indicate disruptions in semantic relationships and local geometry.

Together, these metrics allow us to capture both bias reduction and the preservation (or degradation) of semantic structure.

\section{Experiments and Results}

\subsection{Experimental Setup}

We examine PCA-based gender debiasing on three static word embeddings: GloVe, Word2Vec, and FastText. For each embedding, we construct a gender subspace using PCA over definitional gender pairs and study two experimental settings: cumulative multi-PC removal and single-PC ablation.

In the multi-PC setting, we remove the first $k$ principal components for increasing values of $k$ and evaluate the resulting embedding using direct bias, WEAT, mean vector displacement, and semantic stability. In the single-PC ablation setting, we remove each principal component individually in order to understand how bias and geometry are distributed across the spectrum.

This setup allows us to compare the behavior of direct and associative bias while also quantifying the geometric cost of debiasing.

\subsection{Spectral Structure of the Bias Subspace}

To better understand the structure of the gender subspace, we analyze the explained variance spectrum obtained from PCA over definitional gender pair differences.

\begin{figure}[t]
    \centering
    \includegraphics[width=\linewidth]{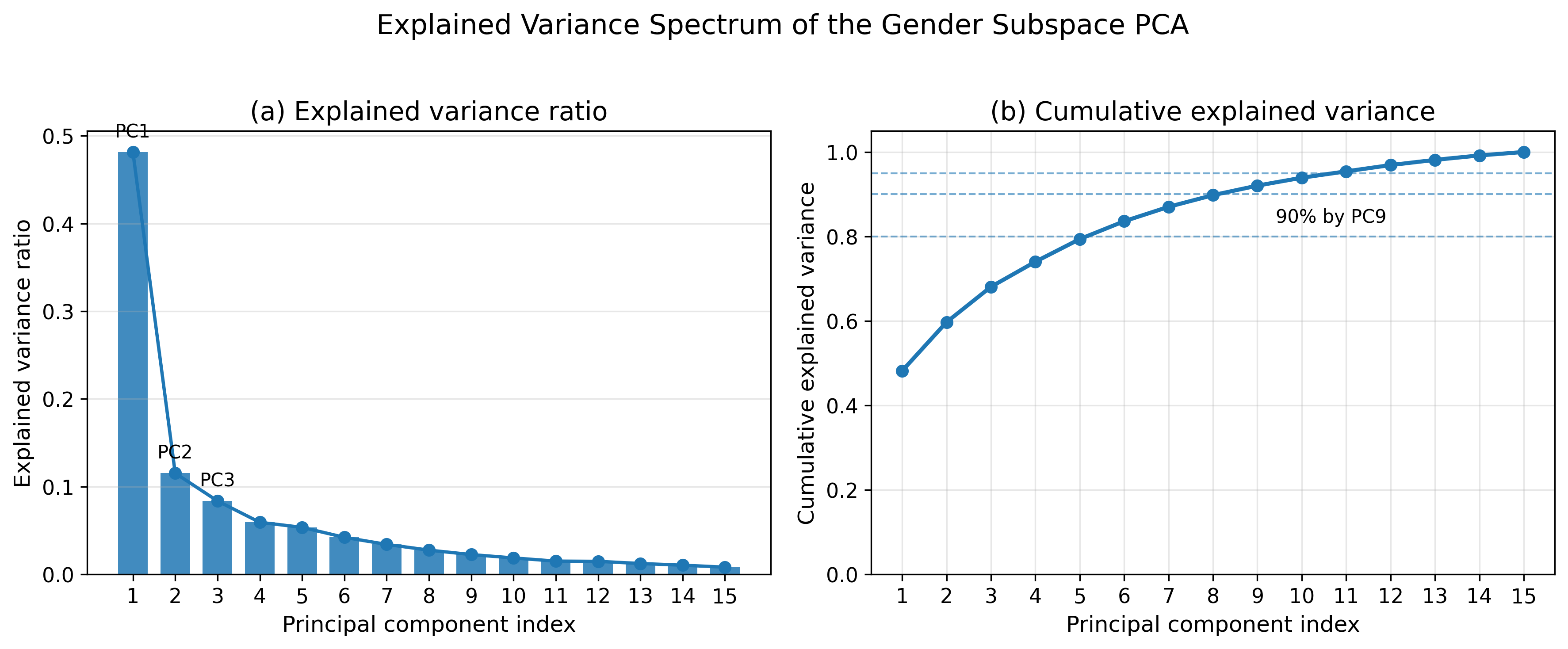}
    \caption{
    Explained variance spectrum of the gender subspace.
    (a) The explained variance ratio of individual principal components shows that the first component captures a substantial portion of the variance.
    (b) The cumulative explained variance indicates that most of the gender-related variation is concentrated in a small number of components, with approximately 90\% of the variance captured within the first 8--10 principal components.
    }
    \label{fig:explained_variance}
\end{figure}

The results reveal a strong concentration of variance in the leading principal components. In particular, the first principal component alone captures approximately $48\%$ of the total variance, while the first five components explain nearly $80\%$, and the first ten exceed $93\%$.

This spectral structure provides direct empirical evidence that the gender bias subspace is inherently low-dimensional. The dominance of the first principal component explains why removing a single direction is sufficient to eliminate most projection-based direct bias, as observed in the multi-PC removal experiments.

At the same time, variance continues to spread across additional components, suggesting that bias is not limited to a single direction. This explains why associative bias, measured by WEAT, does not disappear after removing the first component but remains distributed across multiple dimensions.

\subsection{Cross-Embedding Analysis of Multi-PC Removal}

\begin{figure}[t]
    \centering
    \includegraphics[width=\linewidth]{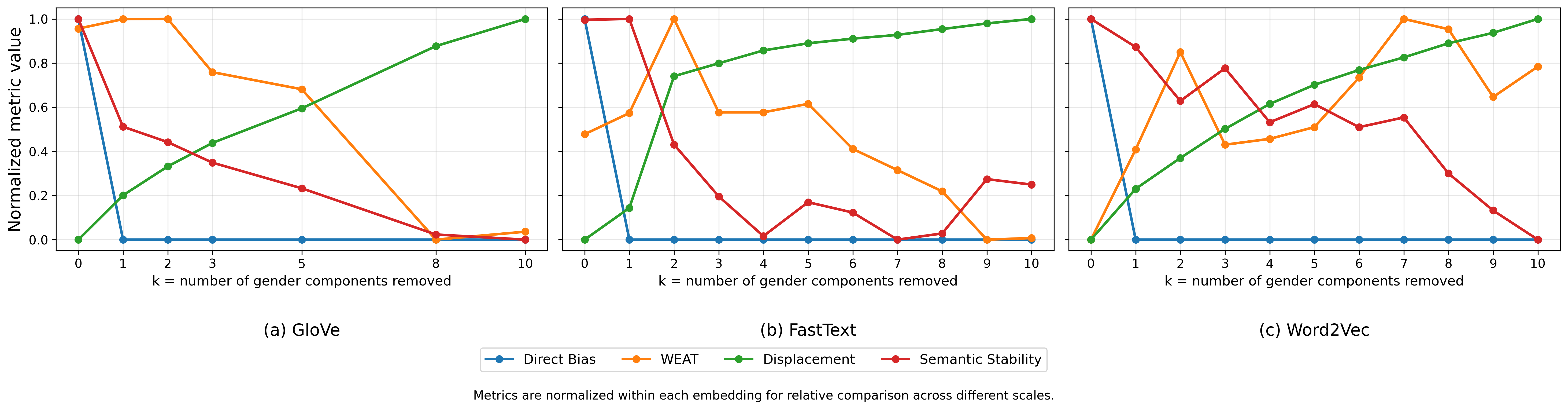}
    \caption{
    Cross-embedding comparison of cumulative PCA-based component removal on GloVe, FastText, and Word2Vec. Direct bias collapses after removal of the first principal component across all embeddings, while WEAT exhibits non-monotonic and embedding-dependent behavior. Mean vector displacement increases steadily and semantic stability declines, demonstrating a trade-off between bias reduction and preservation of embedding geometry.
    }
    \label{fig:cross_embedding_combined}
\end{figure}

We first compare cumulative principal component removal across GloVe, FastText, and Word2Vec. In this setting, removing $k$ components corresponds to removing the first $k$ principal components jointly (e.g., $k=3$ removes components 1, 2, and 3). Figure~\ref{fig:cross_embedding_combined} shows the normalized behavior of direct bias, WEAT, mean vector displacement, and semantic stability as increasing numbers of principal components are removed.

Across all three embeddings, direct bias drops sharply after removal of the first principal component and remains close to zero afterward. This indicates that projection-based direct gender bias is concentrated in a very low-dimensional subspace and is dominated by the leading principal direction.

In contrast, WEAT behaves differently. Its response to cumulative component removal is slower and less consistent, with noticeable non-monotonic variation across embeddings. This suggests that associative bias is not aligned with the same principal directions and remains distributed across multiple dimensions of the embedding space.

At the same time, geometric distortion increases steadily as more components are removed. Mean vector displacement grows monotonically, while semantic stability decreases overall, showing that cumulative projection increasingly alters the structure of the embedding space. Taken together, these results reveal a clear trade-off: direct bias can be reduced quickly, but semantic geometry is progressively degraded.

\subsection{Detailed Multi-PC Behavior}

A more detailed analysis of the cumulative removal process further supports this interpretation. Direct bias decreases sharply after removing the first component and remains nearly constant thereafter, confirming that most projection-based bias is captured by the leading direction.

In contrast, WEAT decreases more gradually and exhibits non-monotonic behavior, indicating that associative bias is not confined to a single direction. At the same time, mean vector displacement increases steadily, and semantic stability decreases, showing that each additional removed component contributes to geometric distortion.

These results demonstrate that while the first component is sufficient to remove direct bias, further removal primarily affects semantic structure rather than providing substantial additional bias reduction.

\subsection{Single-PC Ablation}
To isolate the contribution of individual principal components, we perform a single-PC ablation study in which each component is removed independently. In contrast, the cross-embedding analysis considers cumulative removal, where for a given $k$, the first $k$ principal components are removed jointly (e.g., $k=3$ corresponds to removing components 1, 2, and 3). In the single-PC setting, $k$ instead denotes the index of the removed component, so $k=3$ corresponds to removing only the third principal component.

Figure~\ref{fig:single_pc_normalized} shows that removing the first principal component produces the strongest reduction in direct bias, while removing higher-order components has negligible effect. This confirms that direct bias is highly concentrated in the leading direction.

In contrast, WEAT remains relatively stable across individual ablations and does not show a strong response to any single component. This indicates that associative bias is not localized in one dominant direction, but is instead distributed across multiple dimensions of the embedding space.

The geometric metrics exhibit a different pattern. Both mean vector displacement and semantic stability vary across multiple components, showing that semantic structure is more broadly distributed. While the first component produces the largest geometric change, higher-order components still contribute to non-trivial structural variation.

\subsection{Cross-Embedding Consistency of Single-PC Ablation}
We observe the same qualitative behavior for Word2Vec and FastText. In both embeddings, removing the first principal component reduces direct bias to nearly zero, while removing higher-order components has negligible additional effect. In contrast, WEAT remains relatively stable and exhibits non-monotonic variation across individual ablations, indicating that associative bias is not localized in a single direction. Geometric metrics vary across multiple components, confirming that semantic structure is more broadly distributed. This consistency across embeddings suggests that the low-rank structure of direct bias is a general property of static word embeddings rather than an artifact of a specific model.

\begin{center}
\begin{minipage}{0.95\linewidth}
    \centering
    \includegraphics[width=\linewidth]{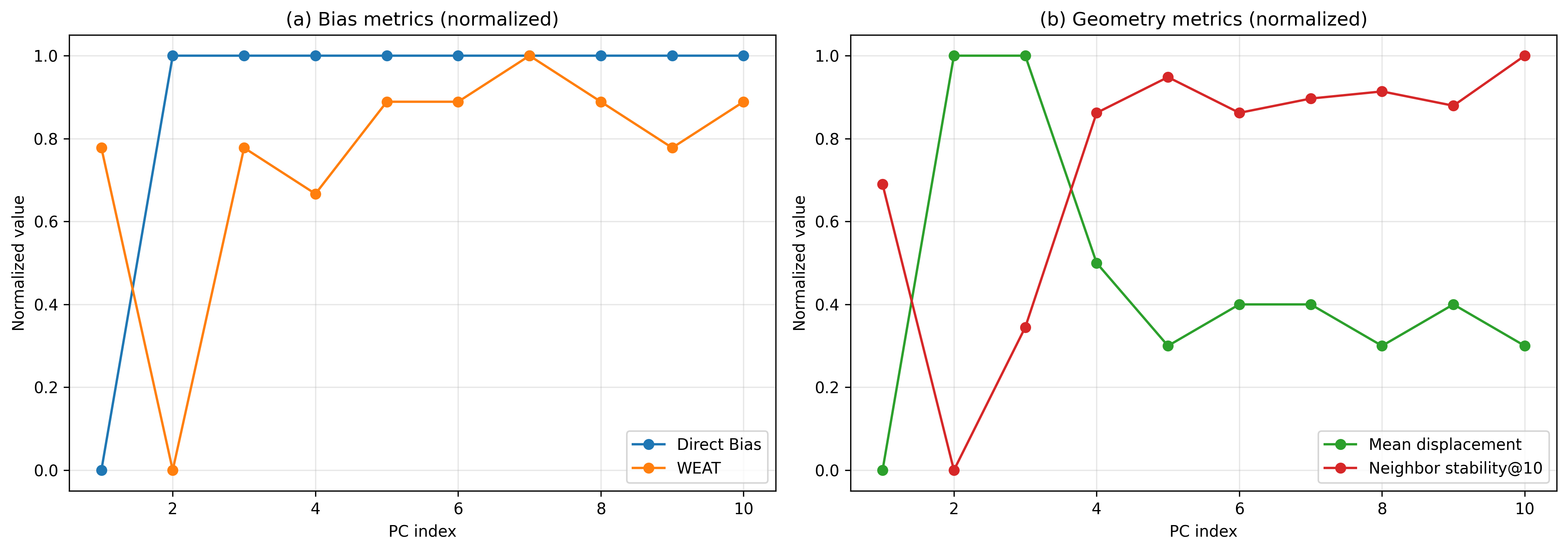}
    \captionof{figure}{
    Single-PC ablation analysis with normalized metrics. (a) Bias metrics (direct bias and WEAT) and (b) geometry metrics (mean vector displacement and semantic stability) are shown as a function of principal component index. Direct bias is dominated by the first principal component, while geometric effects remain distributed across multiple components.
    }
    \label{fig:single_pc_normalized}
\end{minipage}
\end{center}

Overall, these results reinforce the distinction between low-rank direct bias and distributed semantic geometry.

\subsection{Main Takeaways}

Across all experiments, three consistent patterns emerge. First, direct gender bias is strongly concentrated in the first principal component, supporting the low-rank bias hypothesis. Second, associative bias, as measured by WEAT, does not align with these principal directions and remains distributed across multiple dimensions of the embedding space. Third, removing increasing numbers of principal components steadily distorts semantic geometry, producing a clear trade-off between bias reduction and semantic preservation.

These findings indicate that PCA-based debiasing is highly effective for removing projection-based direct bias but does not fully eliminate broader associative bias. At the same time, stronger debiasing interventions increasingly degrade the geometric structure of the embedding space.

\section{Discussion}

Our results provide a consistent geometric interpretation of PCA-based debiasing across multiple word embeddings.

First, the explained variance analysis and single-PC ablation experiments jointly show that direct gender bias is highly concentrated in a low-dimensional subspace, with the first principal component capturing the majority of projection-based bias. This explains why removing a single direction is sufficient to eliminate most direct bias.

However, associative bias behaves fundamentally differently. As demonstrated by WEAT, it does not align with the dominant principal directions and instead remains distributed across multiple dimensions of the embedding space. This suggests that associative bias cannot be effectively removed through low-rank projection alone.

Second, our experiments reveal a clear trade-off between bias reduction and geometric preservation. While removing additional principal components leads to further reductions in certain bias measures, it also introduces increasing distortion in the embedding geometry. This is reflected in growing vector displacement and declining neighborhood stability, indicating that semantic relationships are progressively disrupted.

This trade-off was cross-verified across multiple embeddings such as:  GloVe, FastText, and Word2Vec. Which means that this behavior is not specific to a particular embedding model. It reflects a more general property of distributional representations.

Our findings indicate that PCA-based debiasing is effective for reducing direct bias. Yet insufficient for eliminating broader associative bias. At the same time, aggressive debiasing strategies may compromise the semantic integrity of the embedding space. This highlights the need for more nuanced approaches that can address distributed bias while preserving geometric structure.

\section{Conclusion}

In this work, we presented a geometric analysis of PCA-based gender debiasing in word embeddings. We showed that direct gender bias is strongly concentrated in the leading principal component, supporting the low-rank bias hypothesis. At the same time, associative bias remains distributed across multiple dimensions and does not align with the dominant principal directions.

We further demonstrated that removing increasing numbers of principal components introduces a consistent degradation of embedding geometry, leading to a trade-off between bias reduction and semantic preservation.

Our findings demonstrate that while PCA-based methods are effective for removing direct bias, they are insufficient to solve the full complexity of bias in word embeddings. Future work should focus on developing methods that can capture and remove distributed bias without compromising the underlying semantic structure.

\paragraph{Code Availability:}
The source code, experimental scripts, and figure generation pipeline used in this study are publicly available at: \url{https://github.com/AlexeyKresin/embedding-bias-geometry}

\section*{Acknowledgments}

The author would like to thank Dr. Jason Rafe Miller, Associate Professor of Computer Science at Hood College, for his mentorship, encouragement, and support throughout this research. This work originated from ideas developed during his graduate-level Artificial Intelligence course and was further refined through independent research under his guidance. His insightful discussions, constructive feedback, and commitment to student research were instrumental in the completion of this study and its presentation at the Hood College SPIRE 2026 research symposium.

\end{document}